\documentclass[runningheads]{MICCAI2026-Latex-Template/llncs}
\usepackage[T1]{fontenc}
\usepackage{graphicx,verbatim}
\usepackage{amssymb}
\usepackage{multirow}
\begin{document}
\title{Refining 3D Medical Segmentation\\with Verbal Instruction}
\titlerunning{Abbreviated paper title}
%

\author{Kangxian Xie\inst{1} \and
Jiancheng Yang\inst{2,3} \and
Nandor Pinter\inst{1}\and
Chao Wu\inst{1}\and
Behzad Bozorgtabar\inst{4}\and
Mingchen Gao\inst{1}}
\authorrunning{K. Xie et al.}
%
\institute{University at Buffalo, Buffalo NY 14260, USA \\
\and
ELLIS Institute Finland, Espoo 02150, Finland\and
Aalto University, Espoo 02150, Finland \\ \and
Aarhus University, Aarhus 8000, Denmark \\
}

  
\maketitle              
\begin{abstract}
Accurate 3D anatomical segmentation is essential for clinical diagnosis and surgical planning. However,  automated models frequently generate suboptimal shape predictions due to factors such as limited and imbalanced training data, inadequate labeling quality, and distribution shifts between training and deployment settings. A natural solution is to iteratively refine the predicted shape based on the radiologists' verbal instructions. However, this is hindered by the scarcity of paired data that explicitly links erroneous shapes to corresponding corrective instructions. As an initial step toward addressing this limitation, we introduce \textbf{CoWTalk}, a benchmark comprising 3D arterial anatomies with controllable synthesized anatomical errors and their corresponding repairing instructions. Building on this benchmark, we further propose an iterative refinement model that represents 3D shapes as vector sets and interacts with textual instructions to progressively update the target shape. Experimental results demonstrate that our method achieves significant improvements over corrupted inputs and competitive baselines, highlighting the feasibility of language-driven clinician-in-the-loop refinement for 3D medical shapes modeling.


\end{abstract}
\section{Introduction}

Accurate 3D reconstruction of anatomical shapes from medical images is essential to diagnosis, measurement, and intervention planning.
Despite the strong progress of modern segmentation and reconstruction networks, radiologists still frequently encounter suboptimal 3D segmentation shapes in practice, due to domain variations (e.g., different scanners, protocols, and populations), limited and imbalanced training data, or inaccurate annotations.

However, radiologists typically have a clear expectation of what the underlying anatomy looks like while inspecting the 3D image volume. They naturally compare the expected shape with the predicted shape and can articulate the differences (e.g., the lumen is disconnected near the proximal end). Motivated by this observation, we address the segmentation problem from a new perspective by iteratively refining the suboptimal shape according to clinicians' instructions.

Research in text-guided shape editing is largely constrained by the availability of datasets like ShapeTalk~\cite{achlioptas2023shapetalk} and ShapeWalk~\cite{Slim_2024_CVPR} that align linguistic context with geometric changes. While ShapeTalk utilizes manual descriptions of shape differences, and ShapeWalk~\cite{Slim_2024_CVPR} relies on CAD model interpolations for fine-grained control, the medical domain remains underexplored because there is no publicly available 3D dataset of paired medical shapes with natural language descriptions for geometric refinement. We bridge this gap by establishing the \textbf{CoWTalk} benchmark on the arterial segments mask of TopCoW~\cite{yang2023benchmarking} dataset. We simulate controlled segmentation errors by introducing parameterized morphological perturbations to the ground truth segments (Fig.~\ref{fig:endpoint_process}), and subsequently associate them with LLM-generated corrective instructions (Fig.~\ref{fig:instruction_generation}). For \textbf{CoWTalk}, we generate 15 erroneous variants for each subject in TopCoW, and construct an (Erroneous shape, Instruction, GT Shape) tuple per variant, resulting in a paired dataset comprising 3,750 samples.

Early research on text-shape interaction primarily focused on language-based shape generation \cite{jain2022zero_dreamfield,wang2022clip_nerf}, texture editing \cite{michel2022text2mesh} based on CLIP \cite{radford2021learning}, and shape editing (ShapeTalk~\cite{achlioptas2023shapetalk}, ShapeWalk \cite{Slim_2024_CVPR}), with the latter two being the most relevant to ours. While ShapeTalk aligns text and geometry via feature concatenation, ShapeWalk adopts a latent diffusion-based mechanism. Implicit-based approach~\cite{berzins2023neural} leverages boundary sensitivity to link neural parameters to surface update. Interactive paradigms such as PrEditor3D~\cite{Erkoc_2025_CVPR} explore synchronized 2D multi-view editing. In contrast, the proposed architecture adopts an intuitive encode-fusion-decode structure for shape refinement. First, the medical structure is encoded into vector sets~\cite{zhang20233dshape2vecset}, and the vector sets undergo multi-modal fusion with text features~\cite{devlin2019bert} before implicit decoding. This pipeline translates high-level verbal instructions into 3D shape edits by conditioning shape representations on the contextual information provided by radiologists.

\begin{figure}[t]
    \centering
    \includegraphics[width=\textwidth]{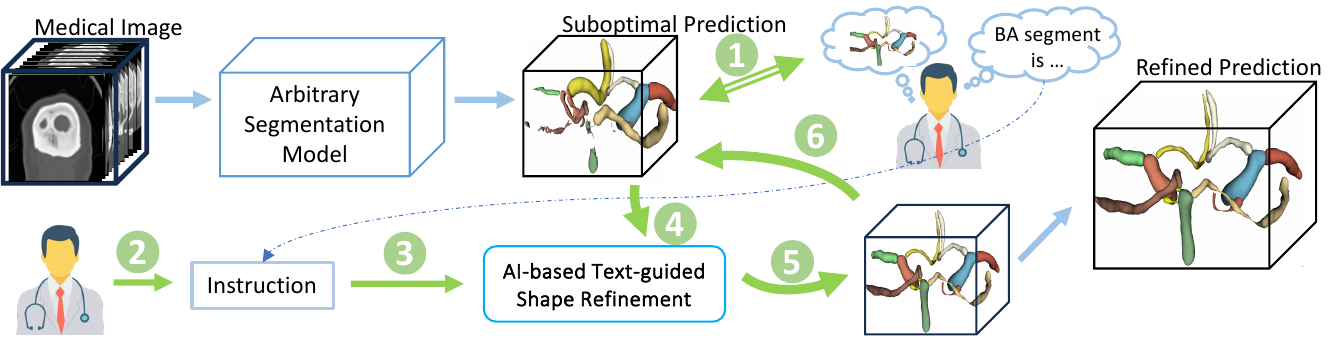}
    \caption{The overall workflow of iterative text-based refinement for 3D medical segmentation shapes. The green arrows and 6 steps represent the iterative workflow.}
    \label{fig:workflow}
\end{figure}

Our contributions are threefold. First, we introduce \textbf{CoWTalk}, a new benchmark for instruction-following medical shape refinement based on TopCoW~\cite{yang2023benchmarking}. Second, we propose an end-to-end 3D vision-language iterative refinement framework~(Fig.~\ref{fig:workflow}) that updates a multi-class 3D segmentation shape according to clinician instructions, enabling human-in-the-loop correction of complex vascular anatomy. Finally, we conduct a comprehensive evaluation to validate the advantages and effectiveness of the proposed method.

\section{Problem Formulation}

There are two stages to achieving instruction-guided medical shape refinement. The first stage synthesizes errors into ground truth~(GT) anatomical shapes and pairs each corrupted shape with a natural-language corrective instruction. The second stage establishes the learning objective, where the model learns to recover the GT shape from a corrupted version given natural language instructions.

\subsection{CoWTalk Dataset Synthesis}
\begin{figure}[t]
    \centering
    \includegraphics[width=\textwidth]{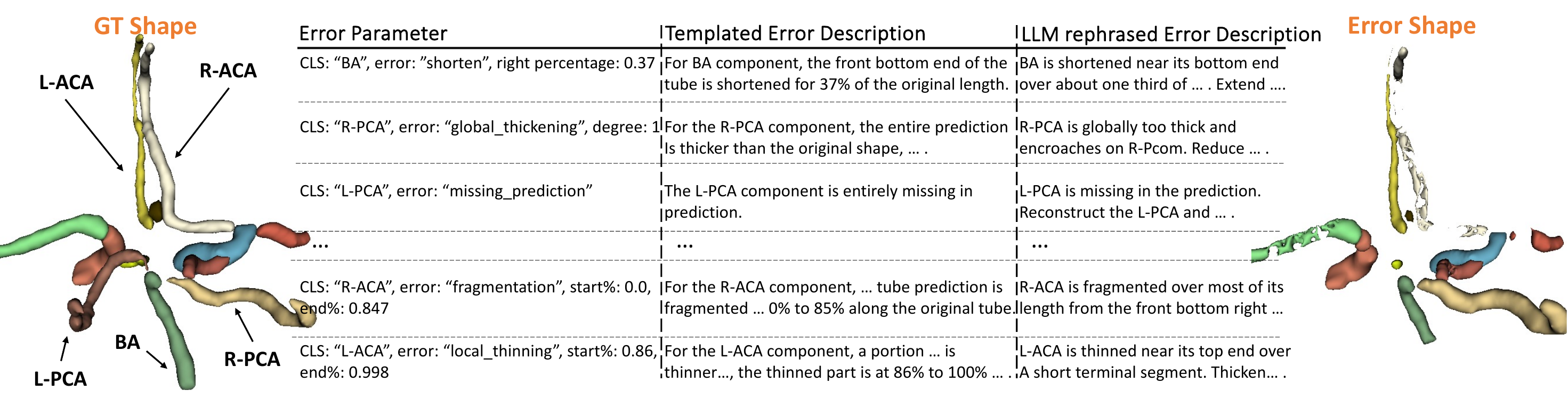}
    \caption{Examples of generated instructions in \textbf{CoWTalk}.}
    \label{fig:instruction_generation}
\end{figure}

The \textbf{CoWTalk} benchmark is developed based on the artery label masks of the Circle of Willis~(CoW) from the TopCoW dataset~\cite{yang2023benchmarking}, which provides the 14-class masks for the artery structure. For each subject, we iterate through its 13 arterial tubular segments, and synthesize error into the segment. The errors are randomized from a designated list of global errors: global thickening/thinning, missing segments, and local errors: local thickening/thinning, shortening, disconnection, fragmentation. For global errors, 3D morphological operations~(dilation, erosion and etc.)~\cite{2020SciPy-NMeth} are applied to the entire segment, while a local error affects only a contiguous portion of the tube. With each error type, error-related parameters~(Fig.\ref{fig:instruction_generation}, left) are also randomly generated as a numerical description. 

\begin{figure}[t]
    \centering
    \includegraphics[width=\textwidth]{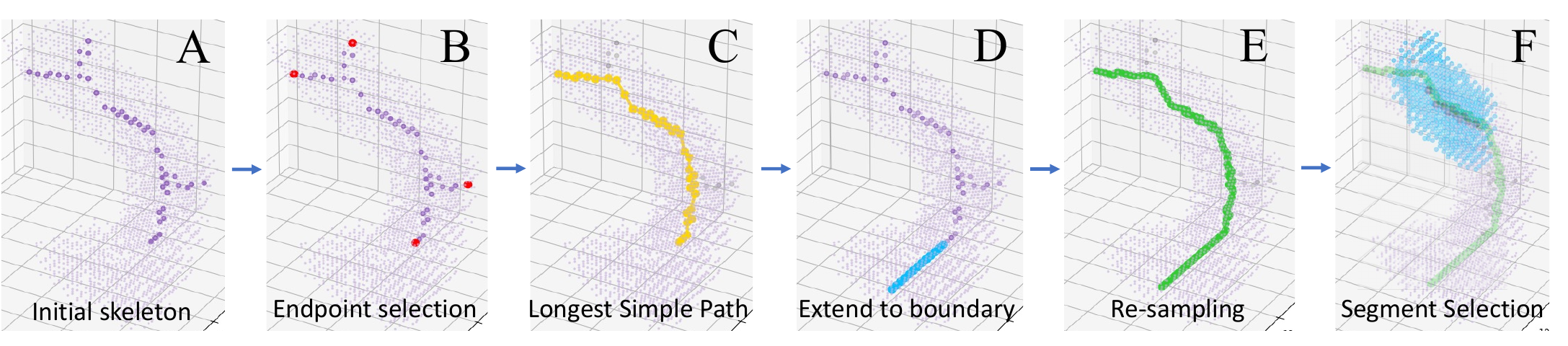}
    \caption{Centerline refinement and Segment Selection. A to E shows the steps for refining the centerline representation, while F illustrates the partial selection of a segment.}
    \label{fig:endpoint_process}
\end{figure}

For local error synthesis, selecting a local tube span from error parameters for error synthesis requires special attention. We use each class segments' skeletal centerlines from the TopCoW centerline dataset~\cite{musio2025circle} and improve upon~(Fig.\ref{fig:endpoint_process} A-E) them to be the abstract spatial and trajectory representation of the shape for local region identification. Given the percentage range parameters~(Fig.\ref{fig:instruction_generation}, left), we treat the centerline points as connected graph nodes with edge weights $1$, and select the points located within the desired range from an anchor endpoint. For example, a $86\%$-to-$99\%$ tube local selection corresponds to selecting the $86$ to $99$ percentile points in terms of their distance away from an anchor endpoint. For each selected point, we generate a cylinder shape pointing at the local trajectory of the centerline. The selected mask~(Fig.\ref{fig:endpoint_process} F) is identified as the intersection between these cylinders and the original ground truth mask. 

As shown in Fig.~\ref{fig:instruction_generation}, given error parameters, we instantiate a template sentence that describes the error, and fix a canonical posterior viewing direction so that directional phrases (e.g., left/right, proximal/distal) are unambiguous. Subsequently, a dedicated prompt is created for ChatGPT 5.2~\cite{openai_api_2026} to paraphrase the templated error description into radiologist-style narratives, followed by corrective instructions with two variants: a concise (general) and a comprehensive (detailed) instruction. 

\subsection{Interactive Shape Refinement Objective}
The refinement objective is to learn a mapping $f_{\theta}$ that iteratively recovers the ground-truth anatomy $S_{gt}$ from an erroneous input shape by resolving the errors specified in a sequence of natural-language instructions $\{t^{k}\}_{k=0}^{K-1}$.
We formulate this as an iterative process:
\begin{equation}
\label{equ:refine}
S^{k+1} = f_{\theta}(S^{k}, t^{k}), \quad k=0,\dots,K-1,
\end{equation}
where $S^{0}$ is the initial erroneous shape and $S^{K}$ is the final refined output, expected to match the ground truth on the targeted arterial structures.

\section{Methods}

\begin{figure}[t]
    \centering
    \includegraphics[width=\textwidth]{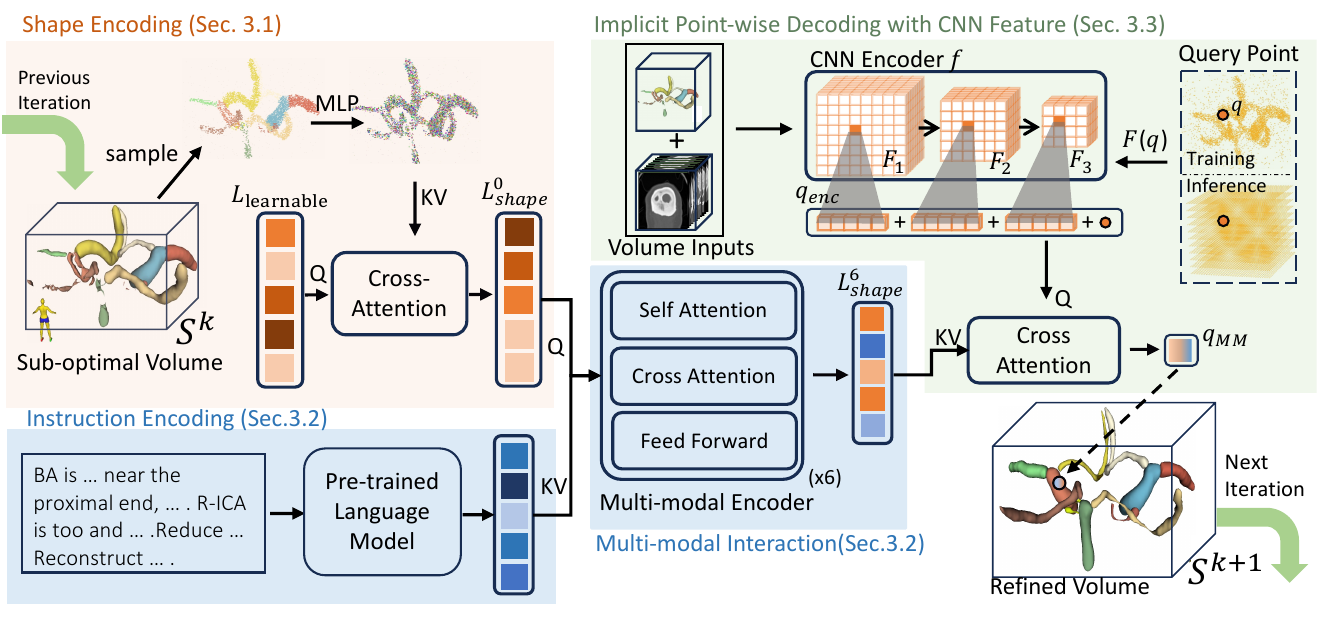}
    \caption{Overview of the proposed instruction-guided medical shape refinement pipeline.}
    \label{fig:arch}
\end{figure}

We propose a language-guided end-to-end shape refinement network~(Fig.\ref{fig:arch}) that takes a suboptimal segmentation shape and the radiologist's instruction as input and outputs a refined segmentation shape according to Equation.~\ref{equ:refine}. The proposed architecture contains four major components: shape encoding, text encoding, multi-modal interaction, and implicit point-wise decoding with CNN features.

\subsection{Shape Encoding via Latent Query Cross-Attention}
Given an input volume containing the corrupted segmentation shape $S_{error}$, we sample a point cloud $P=\{(\mathbf{x}_i,\mathbf{p}_i)\}_{i=1}^{N}$, where $\mathbf{x}_i\in\mathbb{R}^{3}$ denotes the coordinate and $\mathbf{p}_i$ is the label at $\mathbf{x}_i$ in the corrupted volume. 
Inspired by 3DS2VS~\cite{zhang20233dshape2vecset}, we first project the input points into point-wise embedding features $\phi(P)$, and embed the point features $\phi(P)$ into a learnable latent vector set $L_{learnable}\in\mathbb{R}^{M\times d}$ with a cross-attention~\cite{vaswani2017attention} module to summarize the shape:
\begin{equation}
L_{shape} = \mathrm{CrossAttn}(L_{learnable},\, \phi(P)),
\end{equation}
where $L_{shape}\in\mathbb{R}^{M\times d}$~($M=512$, $d=512$) is referred to as shape latents. 
\subsection{Instruction Encoding and Multi-modal Interaction}
A radiologist provides a plain-text instruction $t$ describing what is wrong and how to correct it.
We tokenize $t$ and encode it with a pre-trained BERT encoder~\cite{devlin2019bert} to obtain per-token features in $\mathbb{R}^{768}$, then project token-wise to $d=512$ to form a text feature vector set $T\in\mathbb{R}^{L\times 512}$. We refer to $T$ as the \emph{text feature}.

To inject text feature into the \emph{Shape Latents} $L_{shape}$, we employ a multi-modal encoder consisting of $K=6$ attention blocks.
Each block contains: a self-attention~\cite{vaswani2017attention} layer over the current $L_{shape}$, a cross-attention layer where $L_{shape}$ attends to text features, and a feed-forward network.
Let $L_{shape}^{k}$ denote the shape latent after the $k$-th block, we compute
\begin{equation}
L_{shape}^{k+1} = \mathrm{FFN}\big(\mathrm{CrossAttn}(\mathrm{SelfAttn}(L_{shape}^{k}),\, T)\big), \quad k=0,\dots,K-1.
\end{equation}
The output $L_{shape}^{K}$ is a set of \emph{Shape Latents} conditioned on the user instruction.\\

\subsection{Implicit Point-wise Decoding with CNN Feature}
\label{subsec:decode}

Arterial geometries are locally smooth and near-linear, following well-defined trajectories in 3D space. Implicit functions are well-known to naturally preserve continuity~\cite{berzins2023neural,Park_2019_CVPR,10.1007/978-3-031-16440-8_48,mescheder2019occupancy,NEURIPS2020_53c04118_siran}, but they still require spatially coherent context. To generate such a continuous feature field, we feed the error volume and the image~(optional) through a 3-layer CNN $f$ to obtain feature pyramids $F = \{F_{1},F_{2},F_{3}\}$. For each query location $\mathbf{q}$, we interpolate $F$ at $\mathbf{q}$ for multi-level features, and form the point encoding: $\mathbf{q}_{enc}$ = [$F_{1}(\mathbf{q}_m)$, $F_{2}(\mathbf{q}_m)$, $F_{3}(\mathbf{q})$, $\mathbf{q}$, $S_{error}(\mathbf{q})$]. Then, an cross-attention-based implicit-function decodes $\mathbf{q}_{enc}$ based on the \emph{Shape Latents} $L_{shape}^{K}$ for an updated point encoding with multi-modal context:
\begin{equation}
\mathbf{q}_{MM} = \mathrm{CrossAttn}(\mathbf{q}_{enc}, L_{shape}^{K})
\end{equation}
We then apply an MLP to predict class logits $\hat{\mathbf{y}}_m = \mathrm{MLP}(\mathbf{q}_{MM})$.
Finally, the predicted labels are assembled into the refined segmentation ${S}_{refined}$.

\subsection{Training and Inference}
During training, query points are sampled from the near-surface region, foreground, and random background to ensure sufficient coverage of key locations. We mix the long and short versions of instruction during training. The model is trained end-to-end with CE loss for 200 epochs. During inference, all 3D locations are fed as query points to enable full-volume reconstruction. 

As a training data augmentation, we randomly remove each error instance with probability $0.2$ (i.e., replace it with the correct class shape) as model input. For each input error, the model learns to correct the error with probability $0.5$, guided by the corresponding instruction, leaving the remaining error intact.

\section{Experiments}

We perform experiments against baselines (Sec.\ref{subsec:baseline}), analyze robustness across error types and input modalities~(Sec.~\ref{subsec:input_ablation}), and evaluate the network on real errors with instructions by a radiologist~(Sec.~\ref{subsec:real_ins}). In primary experiments~(Tab.\ref{tab:main_results}) and error type analysis~(Fig.\ref{fig:plot}) on \textbf{CoWTalk}, we drop each error with probability $20\%$ to simulate variable input segmentation quality. 

\subsection{Baseline Comparisons}
\label{subsec:baseline}
Table~\ref{tab:main_results} reports the results for erroneous shape refinement. We compare our method against CNN-based refinement, 3D reconstruction $+$ text refinement, and specialized text-guided shape editing methods in general-purpose domain.

\begin{figure}[t]
\centering
\includegraphics[width=\textwidth]{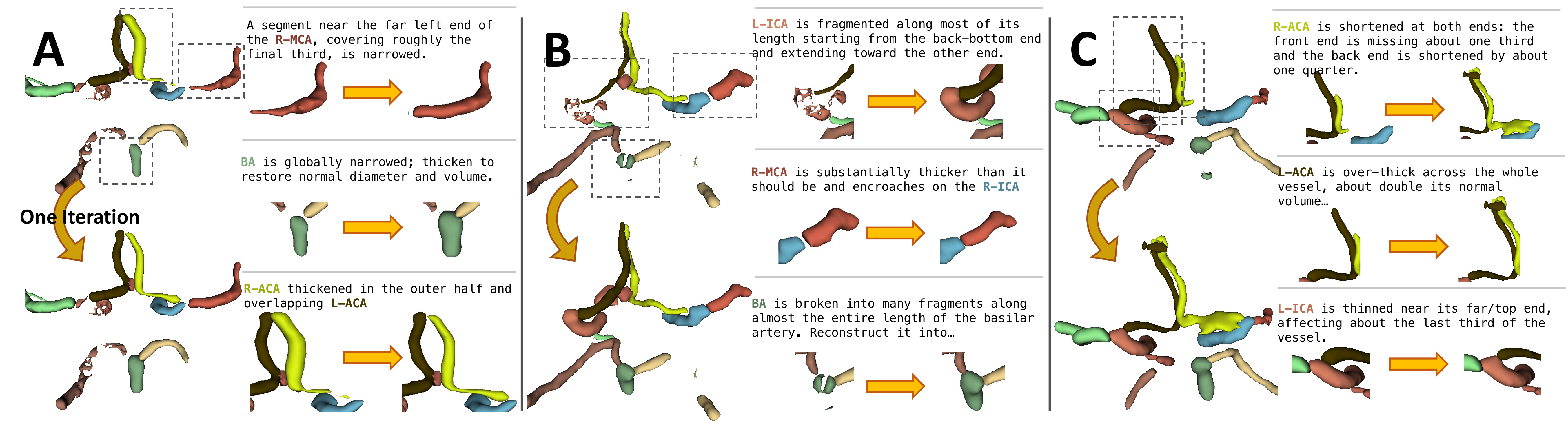}
    \caption{Qualitative refinement results with shape-only input from \textbf{CoWTalk}, and only 3 corrections are highlighted for each sample. Some correction descriptions are omitted.}
\label{fig:qualitative}
\end{figure}

For CNN-based refinement, we tested nnUNetv2~\cite{isensee2024nnu} and SwinUNetrv2~\cite{10.1007/978-3-031-43901-8_40} on two input settings: shape-only input, and 2-channel input with concatenated medical image. We observe that with the image, nnUNetv2 achieves the highest F1, and improves upon the erroneous shape, likely due to its reliance on image.

\begin{table}[t]
\caption{\textbf{CoWTalk} refinement performances using F1 score, Dice, Normalized Surface Dice~(NSD) and Chamfer's Distance~(CD) as metrics. Results are presented both without and with images. Iter.: Iterative; Text: if guided by text. *: methods with specialized enhancements. }\label{tab:main_results}
\centering
\makebox[\textwidth][c]{%
\begin{tabular}{l c@{\hspace{10pt}}c@{\hspace{12pt}}c@{\hspace{7pt}}c@{\hspace{7pt}}c@{\hspace{7pt}}c}
\hline
Method & Iter. & Text & F1 & Dice & NSD & CD($\downarrow$) \\
\hline
Input Error & -- & -- & 86.0  & 81.9 & 86.7 & 0.795 \\
\hline
nnUNetv2~\cite{isensee2024nnu} & -- & -- & 84.6 / \textbf{94.2} & 79.5 / 90.0 & 86.4 / 93.2 & 1.84 / \textbf{0.29} \\
SwinUnetrv2~\cite{10.1007/978-3-031-43901-8_40} & -- & -- & 86.6 / 89.5 & 79.7 / 84.7 & 84.5 / 90.0 & 0.83 / 0.77 \\
3DS2VS~\cite{zhang20233dshape2vecset} & -- & \checkmark & 39.6 / -- & 36.3 / -- & 34.2 / -- & 5.88 / -- \\
ShapeTalk~\cite{achlioptas2023shapetalk}* & \checkmark & \checkmark & 72.4 / 79.3 & 71.0 / 78.7 & 77.6 / 83.6 & 3.13 / 3.28 \\
ShapeWalk~\cite{Slim_2024_CVPR}* & \checkmark & \checkmark & 71.6 / 77.0 & 68.5 / 74.4 & 75.4 / 81.4 & 5.58 / 3.45 \\
\hline
Ours & \checkmark & \checkmark & \textbf{87.7} / 91.8 & \textbf{87.5} / \textbf{91.8} & \textbf{90.7} / \textbf{94.3} & 1.70 / 1.33 \\
\hline
\end{tabular}%
}
\end{table}

Next, we focus on methods based on the latent vector set representation~\cite{zhang20233dshape2vecset}. For methods using vector sets for representation, we unify the vector dimensionality to $(512,512)$ for fair comparison. We also split the refinement into 4 non-overlapping instruction sets for iterative operation, since the model performs increasingly well as more iterations divide the instruction into shorter text input (Iterations: 1,2,3,4 $\rightarrow$ $91.0, 91.3, 91.6, 91.8$ in Dice \%). First, we evaluate a straightforward baseline that injects textual context via cross-attention into the 3DS2VS~\cite{zhang20233dshape2vecset} reconstruction pipeline to perform edits; however, it fails to reconstruct the original shape. This suggests that na\"{i}vely conditioning generic reconstruction features on text can be unstable, and that the MLP-only implicit decoding feature lacks continuous feature context. We also compare against two general-purpose text-guided shape editing methods, ShapeTalk (S.T.)~\cite{achlioptas2023shapetalk} and ShapeWalk (S.W.)~\cite{Slim_2024_CVPR}. To enable fair comparisons and continuous feature context, we replace the original PointNet~\cite{qi2017pointnet} auto-encoder in S.T. with a 3DS2VS-style vector-set auto-encoder~\cite{zhang20233dshape2vecset}, and further augment both S.T. and S.W. with CNN-derived local features~(Sec.\ref{subsec:decode}) for reconstruction pre-training. Nevertheless, under pretrained-latent-editing protocol, neither method sufficiently reaches comparable refinement performance. Conversely, without medical images as guidance, our text-guided model achieves 6.8\% Dice refinement improvement over the error samples, 10\% advantage over CNN-refinement, and top performances across the first 3 metrics. With image input, our method still outperforms segmentation-focused methods~\cite{isensee2024nnu,10.1007/978-3-031-43901-8_40} in Dice and NSD.

\subsection{Multi-modal Input Ablation on Error Categories}
\label{subsec:input_ablation}
\begin{figure}[t]
    \centering
    \includegraphics[width=\textwidth]{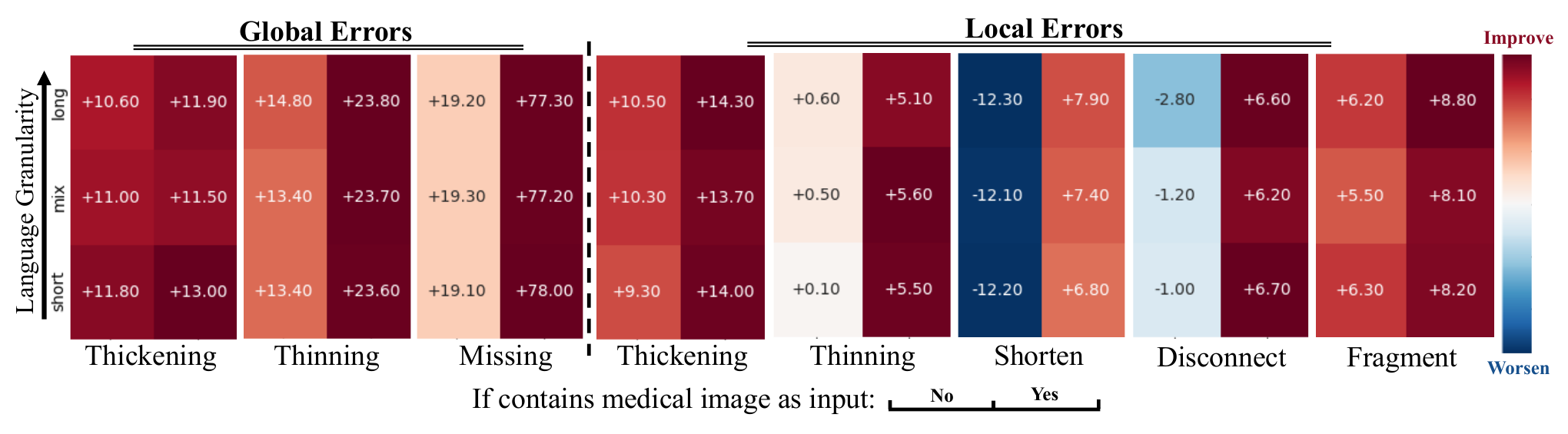}
    \caption{Refinement~(F1 \%) with the proposed method under different visual and language input, showing the difference between refined results and synthesized error~($\pm$).}
    \label{fig:plot}
\end{figure}

We ablate visual and language inputs to quantify the impact of image context and instruction granularity, and display results as heatmaps by error type in Fig.~\ref{fig:plot}. Without image as ground truth, errors that remove anatomy~(Disconnection, Missing, Shorten) are harder to correct as they require fabrication~(Fig.~\ref{fig:qualitative} C. top); in contrast, shape-only refinement works well at errors where topology is largely preserved, such as fragmentation and local thickening~(Fig.~\ref{fig:qualitative} A.~Bottom; B.~Top). Additionally, across most error types, our performance is consistent and robust under various instruction granularities.

\subsection{Radiologist Instruction on Real Error}
\label{subsec:real_ins}
\begin{figure}[t]
    \centering
    \includegraphics[width=\textwidth]{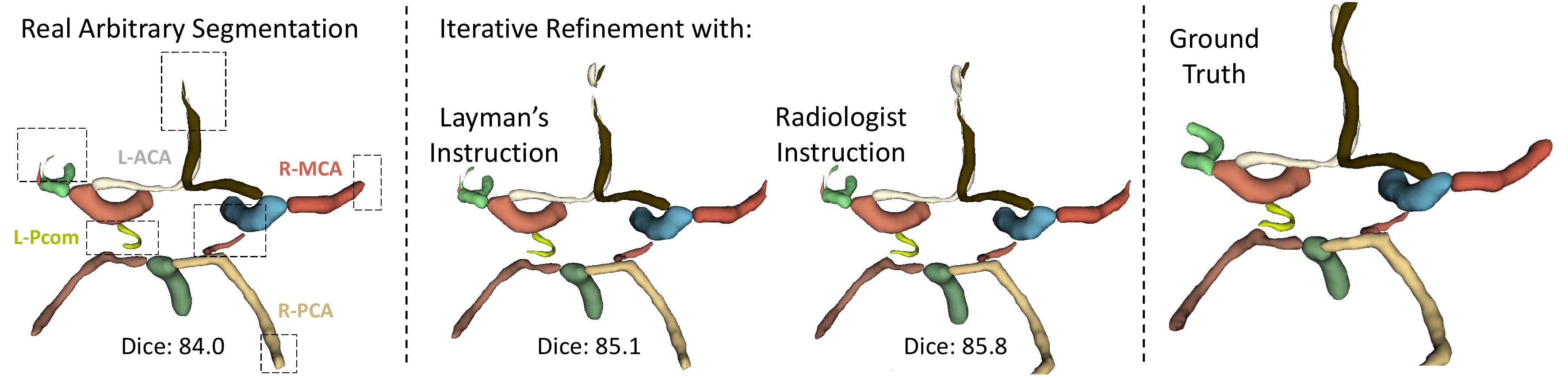}
    \caption{Example of refining a real, suboptimal segmentation~\cite{isensee2024nnu} prediction using instructions given by a layperson and a radiologist.}
    \label{fig:real_sample}
\end{figure}

To assess real-world applicability, we evaluate our method on cases pre-segmented by nnUNetv2~\cite{isensee2024nnu} with instructions provided by a researcher without a medical background, and by a practicing neuroradiologist with 13 years of experience. The qualitative example in Fig.~\ref{fig:real_sample} shows that both instructions correct the error with professional instruction yielding more topologically robust results.  

\section{Conclusion}
This work aims to enable clinician-in-the-loop refinement of medical segmentations through iterative, instruction-following updates. To this end, we introduced a new benchmark \textbf{CoWTalk} by synthesizing controllable vessel-segment errors on TopCoW and pairing them with radiologist-like corrective instructions. An iterative refinement framework is proposed with multi-modal interaction to apply corrective edits. Experiments demonstrate that this pipeline effectively follows instructions and improves segmentation quality.

%
%
%
%
\bibliographystyle{MICCAI2026-Latex-Template/splncs04}
\bibliography{MICCAI2026-Latex-Template/bib}

@article{zhang20233dshape2vecset,
  title={3dshape2vecset: A 3d shape representation for neural fields and generative diffusion models},
  author={Zhang, Biao and Tang, Jiapeng and Niessner, Matthias and Wonka, Peter},
  journal={ACM Transactions On Graphics (TOG)},
  volume={42},
  number={4},
  pages={1--16},
  year={2023},
  publisher={ACM New York, NY, USA}
}

@inproceedings{achlioptas2023shapetalk,
    title={{ShapeTalk}: A Language Dataset and Framework for 3D Shape Edits and Deformations},
    author={Achlioptas, Panos and Huang, Ian and Sung, Minhyuk and Tulyakov, Sergey and Guibas, Leonidas},    
    booktitle={Conference on Computer Vision and Pattern Recognition (CVPR)},    
    year={2023}}

@inproceedings{
berzins2023neural,
title={Neural Implicit Shape Editing using Boundary Sensitivity},
author={Arturs Berzins and Moritz Ibing and Leif Kobbelt},
booktitle={The Eleventh International Conference on Learning Representations },
year={2023},
url={https://openreview.net/forum?id=CMPIBjmhpo}
}

@InProceedings{Slim_2024_CVPR,
    author    = {Slim, Habib and Elhoseiny, Mohamed},
    title     = {ShapeWalk: Compositional Shape Editing Through Language-Guided Chains},
    booktitle = {Proceedings of the IEEE/CVF Conference on Computer Vision and Pattern Recognition (CVPR)},
    month     = {June},
    year      = {2024},
    pages     = {22574-22583}
}

@InProceedings{Erkoc_2025_CVPR,
    author    = {Erko\c{c}, Ziya and G\"umeli, Can and Wang, Chaoyang and Nie{\ss}ner, Matthias and Dai, Angela and Wonka, Peter and Lee, Hsin-Ying and Zhuang, Peiye},
    title     = {PrEditor3D: Fast and Precise 3D Shape Editing},
    booktitle = {Proceedings of the IEEE/CVF Conference on Computer Vision and Pattern Recognition (CVPR)},
    month     = {June},
    year      = {2025},
    pages     = {640-649}
}

@misc{yang2023benchmarking,
      title={Benchmarking the CoW with the TopCoW Challenge: Topology-Aware Anatomical Segmentation of the Circle of Willis for CTA and MRA},
      author={Kaiyuan Yang and Franziska Musio and Yu Ma and Nicola Juchler and Janine Paetzold and Rana Al-Maskari and Konrad Scheffler and Oliver Gross and Robert Tarns Rusu and Ender Konukoglu and Bastian Grange and Hoel Merck and Raphael Sznitman and Roland Wiest and Hongwei Li and Christine A. Chung and Marcel Alexander Schneider and Jens Petersen and Bjoern Menze},
      year={2023},
      eprint={2312.17670},
      archivePrefix={arXiv},
      primaryClass={cs.CV}
}

@article{musio2025circle,
  title={Circle of Willis Centerline Graphs: A Dataset and Baseline Algorithm},
  author={Musio, Fabio and Juchler, Norman and Yang, Kaiyuan and Shit, Suprosanna and Prabhakar, Chinmay and Menze, Bjoern and Hirsch, Sven},
  journal={arXiv preprint arXiv:2510.13720},
  year={2025}
}

@inproceedings{devlin2019bert,
  title={Bert: Pre-training of deep bidirectional transformers for language understanding},
  author={Devlin, Jacob and Chang, Ming-Wei and Lee, Kenton and Toutanova, Kristina},
  booktitle={Proceedings of the 2019 conference of the North American chapter of the association for computational linguistics: human language technologies, volume 1 (long and short papers)},
  pages={4171--4186},
  year={2019}
}

@misc{openai_api_2026,
  author       = {{OpenAI}},
  title        = {OpenAI API},
  year         = {2026},
  howpublished = {\url{https://platform.openai.com}},
  note         = {Accessed February 2026}
}

@inproceedings{isensee2024nnu,
  title={nnu-net revisited: A call for rigorous validation in 3d medical image segmentation},
  author={Isensee, Fabian and Wald, Tassilo and Ulrich, Constantin and Baumgartner, Michael and Roy, Saikat and Maier-Hein, Klaus and Jaeger, Paul F},
  booktitle={International Conference on Medical Image Computing and Computer-Assisted Intervention},
  pages={488--498},
  year={2024},
  organization={Springer}
}

@InProceedings{10.1007/978-3-031-43901-8_40,
author="He, Yufan
and Nath, Vishwesh
and Yang, Dong
and Tang, Yucheng
and Myronenko, Andriy
and Xu, Daguang",
editor="Greenspan, Hayit
and Madabhushi, Anant
and Mousavi, Parvin
and Salcudean, Septimiu
and Duncan, James
and Syeda-Mahmood, Tanveer
and Taylor, Russell",
title="SwinUNETR-V2: Stronger Swin Transformers with Stagewise Convolutions for 3D Medical Image Segmentation",
booktitle="Medical Image Computing and Computer Assisted Intervention -- MICCAI 2023",
year="2023",
publisher="Springer Nature Switzerland",
address="Cham",
pages="416--426",
isbn="978-3-031-43901-8"
}

@inproceedings{qi2017pointnet,
  title={Pointnet: Deep learning on point sets for 3d classification and segmentation},
  author={Qi, Charles R and Su, Hao and Mo, Kaichun and Guibas, Leonidas J},
  booktitle={Proceedings of the IEEE conference on computer vision and pattern recognition},
  pages={652--660},
  year={2017}
}

@InProceedings{10.1007/978-3-031-16440-8_48,
author="Yang, Jiancheng
and Shi, Rui
and Wickramasinghe, Udaranga
and Zhu, Qikui
and Ni, Bingbing
and Fua, Pascal",
editor="Wang, Linwei
and Dou, Qi
and Fletcher, P. Thomas
and Speidel, Stefanie
and Li, Shuo",
title="Neural Annotation Refinement: Development of a New 3D Dataset for Adrenal Gland Analysis",
booktitle="Medical Image Computing and Computer Assisted Intervention -- MICCAI 2022",
year="2022",
publisher="Springer Nature Switzerland",
address="Cham",
pages="503--513",
isbn="978-3-031-16440-8"
}

@InProceedings{Park_2019_CVPR,
author = {Park, Jeong Joon and Florence, Peter and Straub, Julian and Newcombe, Richard and Lovegrove, Steven},
title = {DeepSDF: Learning Continuous Signed Distance Functions for Shape Representation},
booktitle = {The IEEE Conference on Computer Vision and Pattern Recognition (CVPR)},
month = {June},
year = {2019}
}

@inproceedings{michel2022text2mesh,
  title={Text2mesh: Text-driven neural stylization for meshes},
  author={Michel, Oscar and Bar-On, Roi and Liu, Richard and Benaim, Sagie and Hanocka, Rana},
  booktitle={Proceedings of the IEEE/CVF conference on computer vision and pattern recognition},
  pages={13492--13502},
  year={2022}
}

@inproceedings{jain2022zero_dreamfield,
  title={Zero-shot text-guided object generation with dream fields},
  author={Jain, Ajay and Mildenhall, Ben and Barron, Jonathan T and Abbeel, Pieter and Poole, Ben},
  booktitle={Proceedings of the IEEE/CVF conference on computer vision and pattern recognition},
  pages={867--876},
  year={2022}
}

@inproceedings{wang2022clip_nerf,
  title={Clip-nerf: Text-and-image driven manipulation of neural radiance fields},
  author={Wang, Can and Chai, Menglei and He, Mingming and Chen, Dongdong and Liao, Jing},
  booktitle={Proceedings of the IEEE/CVF conference on computer vision and pattern recognition},
  pages={3835--3844},
  year={2022}
}

@inproceedings{mescheder2019occupancy,
  title={Occupancy networks: Learning 3d reconstruction in function space},
  author={Mescheder, Lars and Oechsle, Michael and Niemeyer, Michael and Nowozin, Sebastian and Geiger, Andreas},
  booktitle={Proceedings of the IEEE/CVF conference on computer vision and pattern recognition},
  pages={4460--4470},
  year={2019}
}

@inproceedings{radford2021learning,
  title={Learning transferable visual models from natural language supervision},
  author={Radford, Alec and Kim, Jong Wook and Hallacy, Chris and Ramesh, Aditya and Goh, Gabriel and Agarwal, Sandhini and Sastry, Girish and Askell, Amanda and Mishkin, Pamela and Clark, Jack and others},
  booktitle={International conference on machine learning},
  pages={8748--8763},
  year={2021},
  organization={PmLR}
}

@ARTICLE{2020SciPy-NMeth,
  author  = {Virtanen, Pauli and Gommers, Ralf and Oliphant, Travis E. and
            Haberland, Matt and Reddy, Tyler and Cournapeau, David and
            Burovski, Evgeni and Peterson, Pearu and Weckesser, Warren and
            Bright, Jonathan and {van der Walt}, St{\'e}fan J. and
            Brett, Matthew and Wilson, Joshua and Millman, K. Jarrod and
            Mayorov, Nikolay and Nelson, Andrew R. J. and Jones, Eric and
            Kern, Robert and Larson, Eric and Carey, C J and
            Polat, {\.I}lhan and Feng, Yu and Moore, Eric W. and
            {VanderPlas}, Jake and Laxalde, Denis and Perktold, Josef and
            Cimrman, Robert and Henriksen, Ian and Quintero, E. A. and
            Harris, Charles R. and Archibald, Anne M. and
            Ribeiro, Ant{\^o}nio H. and Pedregosa, Fabian and
            {van Mulbregt}, Paul and {SciPy 1.0 Contributors}},
  title   = {{{SciPy} 1.0: Fundamental Algorithms for Scientific
            Computing in Python}},
  journal = {Nature Methods},
  year    = {2020},
  volume  = {17},
  pages   = {261--272},
  adsurl  = {https://rdcu.be/b08Wh},
  doi     = {10.1038/s41592-019-0686-2},
}

@inproceedings{NEURIPS2020_53c04118_siran,
  author = {Sitzmann, Vincent and Martel, Julien and Bergman, Alexander and Lindell, David and Wetzstein, Gordon},
  booktitle = {Advances in Neural Information Processing Systems},
  editor = {H. Larochelle and M. Ranzato and R. Hadsell and M.F. Balcan and H. Lin},
  pages = {7462--7473},
  publisher = {Curran Associates, Inc.},
  title = {Implicit Neural Representations with Periodic Activation Functions},
  volume = {33},
  year = {2020}
}

@article{vaswani2017attention,
  title={Attention is all you need},
  author={Vaswani, Ashish and Shazeer, Noam and Parmar, Niki and Uszkoreit, Jakob and Jones, Llion and Gomez, Aidan N and Kaiser, {\L}ukasz and Polosukhin, Illia},
  journal={Advances in neural information processing systems},
  volume={30},
  year={2017}
}

\end{document}